# A New Semisupervised Technique for Polarity Analysis using Masked Language Models

Kohei Watanabe

Waseda University and LUT University

Abstract

I developed a new version of Latent Semantic Scaling (LSS) employing word2vec as a masked language model. Unlike original spatial models, it assigns polarity scores to words and documents as predicted probabilities of seed words to occur in given contexts. These probabilistic polarity scores are more accurate, interpretable and consistent than those spatial polarity models can produce in text analysis. I demonstrate these advantages by applying both probabilistic and spatial models to China Daily's coverage of China and other countries during the coronavirus disease (COVID) pandemic in terms of achievement in health issues. The result suggests that more advanced masked language models would further improve the semisupervised machine learning technique.

## Introduction

Polarity is one of the most important measurements in text analysis because it captures the document's emphasis on specific concepts. The types of polarity vary from affective sentiment (Young & Soroka, 2012), political ideology (Slapin & Proksch, 2008) to temporal orientation (Watanabe & Sältzer, 2023), depending on the theoretical framework of research. Polarity analysis can be performed using various techniques but they have strengths and weaknesses: keyword matching works regardless of the size of the corpora but requires a long list of words; supervised machine learning performs well but demands an annotated corpus; unsupervised machine learning is inexpensive but often unsuitable for capturing pre-defined concepts (Watanabe, 2020).

Semisupervised machine learning methods were proposed to make polarity analysis more accessible in languages or domains where appropriate keyword dictionaries or annotated corpora are not available. Latent Semantic Scaling (Watanabe, 2020) achieves polarity analysis by automatically expanding a short list of user-provided seed words and assigning polarity scores to all the words in the target corpus. This technique allows researchers to estimate the polarity scores for all the documents in the corpus on a pre-defined dimension without additional costs.

In fact, LSS has been used successfully by scholars in various fields to analyze textual data from unique angles. For example, Umansky (2022) analyzed social media posts focusing on environmental issues in the Amazon rainforest. Brazys et al. (2023) detected the use of aggressive languages by Chinese diplomats on social media. Eder (2024) measured the influence of populist and radical parties through Austria's foreign policy documents. Yamao (2024) studied reconciliation between Sunni and Shia Muslims in Iraqi newspapers. Ito et al. (2025) revealed Chinese leaders' market economy orientation through their speeches. Seki (2025) studied

Turkey's democratic reforms through the European Union's reports. LSS was also used to assess the robustness of dictionary analysis (Baerg & Krainin, 2022; Wong, 2026).

However, the original LSS model has limitations that prevented it from wider adoption (Mochihashi, 2021). First, the estimated polarity scores for documents tend to have large measurement errors without aggregation by other variables (e.g., dates, authors or topics); second, the polarity scores do not have substantive meanings; third, seed words must have extreme polarity to assign a wide range of scores to documents; fourth, the hyperparameters of the models (e.g., size of word vectors) cannot be optimized easily when singular value decomposition (SVD) is used as the underlying algorithm.

Above problems are inherent to spatial modeling of word semantics where the relevance of words is measured by the cosine similarity of their multidimensional vectors. In this model, user-provided seed words must be placed in extremity because they are the reference points in the semantic space and only words located between them receive meaningful polarity scores. Word vectors are the foundation of spatial modeling, but they are merely low-dimensional representations of the document-term matrix by SVD.

To overcome the limitations, I developed a new version of LSS using word2vec (Mikolov et al., 2013) as a masked-language model.[1] Although word2vec is widely used to train word vectors, the proposed technique employes it as a small language model to compute polarity scores as the probability of the occurrences of seed words. The use of the language model makes the result of LSS more interpretable and consistent with large dictionaries. The model also allows the use of semantically moderate seed words and the optimization of hyperparameters based on perplexity scores.

[1] Mochihashi (2021) proposed a probabilistic version of the technique (called Probabilistic Latent Semantic Scaling) to address these issues, but he used word2vec only for training word vectors.

Among several language models available, word2vec is chosen because its relatively simple algorithm allows us to fully understand the model's functioning, implement it in an open-source software package, train it solely on a local corpus and inspect its parameters for validation. These make researchers accountable for the methodology, ensuring that their analysis is reproducible, transparent and impartial without bias originating from proprietary models (Navigli et al., 2023; Palmer et al., 2024).

In the following sections, I will first explain how LSS is implemented using word2vec focusing on the difference between spatial and probabilistic polarity estimation.[2] While earlier studies used word2vec to compute the spatial similarity between words (e.g., Fu, 2023; Rodman, 2020; Wong, 2026),[3] the current study uses it to predict the probability of words to occur in given contexts. Second, I will evaluate spatial and probabilistic LSS using the corpus of English-language articles published by China's Daily over the coronavirus disease (COVID) pandemic. The evaluation compares document polarity scores on 'achievement' and 'health' between four LSS models and a large dictionary. Finally, I present polarity scores predicted by probabilistic models separately for articles about China and other countries. This serves as an example of research on China's public diplomacy during the public health crisis.

---

[2] The proposed LSS models are implemented in the LSX and wordvector packages and published on CRAN. The word2vec algorithm is implement based on word2vec++ (Fomichev, 2017) but significantly altered to extract output layer weights for probabilistic models.

[3] For methodological discussions on word 2vec, see Borms et al. (2021), Eichstaedt et al. (2021) and Widmann and Wich (2023).

The results show that word2vec-based probabilistic models perform best on both dimensions and that optimized models lead us to the same conclusion as manually complied dictionaries. I argue that the new version of LSS will make semisupervised polarity analysis of media content more accurate, intuitive and objective. This study also demonstrates the possibility to incorporate masked-language models into semisupervised text analysis for further enhancement of the methodology.

## Algorithms

The original LSS is an extension of latent semantic analysis (Deerwester et al., 1990) that applies SVD to reduce the dimension of a sentence-term matrix constructed from a corpus. The resulting word vectors allow LSS to estimate similarity of words based on the contexts of their use and assign polarity scores to all the words in the corpus (Watanabe, 2020). The model is *bipolar* if seed words have their positive and negative polarity scores; it is *unipolar* if they have only positive polarity scores. Unipolar models often produce results comparable to large keyword dictionaries.

Word vectors for LSS can be trained using other algorithms such as glove (Pennington et al., 2014) and word2vec (Mikolov et al., 2013). They often improve the accuracy of LSS, but they are still *spatial* models because polarity scores are based on the similarity of words vectors. Since spatial polarity scores can be interpreted relatively only within the same semantic space, it is hard to compare results from separate models. Further, since spatial polarity scores can take negative values in unipolar models, it is difficult to analyze interaction between multiple variables.[4] These properties limited the usefulness of LSS as an alternative to dictionary analysis.

The proposed LSS technique will solve above problems using word2vec as a language model that predicts the occurrences of words based on their context. In this *probabilistic* model, polarity scores of words are the probability of the seed words to occur in their neighborhood. This approach will enhance the usefulness of LSS in replacing dictionary analysis because probabilistic polarity scores are easier to interpret and compare between models.

**Dictionary Analysis**

Dictionary analysis is old but still a popular method in communication research. Usually, a dictionary comprises words (or their stems) about a target concept manually collected from existing dictionaries, thesauri or corpora (Watanabe & Zhou, 2022). Analysis is performed by matching the dictionary words and normalizing the frequency by the total number of words in each document.

In this simple pattern matching approach, the polarity scores of words are Boolean. The dictionary words are a sample from a population of words about the concept, so the resulting probability $\hat{y}$ is the estimated polarity of the documents:

$$\hat{y} = \frac{1}{N}\sum_{i=1}^{N} g_i f_i$$

[4] When polarity scores are not bounded between 0 and 1, simple products of two lowest (negative) scores result in a higher (positive) score. For the same reason, harmonic means do not allow scores to be negative.

where $f_i$ and $N$ demotes the frequency of dictionary word $i$ and total number of words in the document, respectively. The polarity score $g_i$ is one if word $i$ is among the dictionary words; zero otherwise. To estimate the polarity of documents accurately, a dictionary must include hundreds to thousands of words. If the size of the dictionary is insufficient, the polarity of documents is underestimated.[5]

**Latent Semantic Scaling**

LSS is developed to perform polarity analysis of documents without large dictionaries. It accomplishes analysis comparable to dictionaries' by automatically expanding a small set of seed words. It assigns continuous polarity scores to all the words in the target corpus and uses them to estimate the polarity scores of documents. In this approach, a document polarity score $\hat{y}$ is computed as the weighted average of word polarity scores:

$$\hat{y} = \frac{1}{N}\sum_{i=1}^{N} \hat{g}_i f_i$$

where $\hat{g}_i$ is the score assigned to word $i$ by an algorithm.

The most important step in LSS is the estimation of word polarity scores based on seed words as the only manual inputs. Word polarity scores were cosine similarity to the seed words in spatial models, but they are the likelihoods of the seed words to occur in the same context in probabilistic models. The likelihoods are estimated by a language model.

[5] Underestimation of polarity scores of documents by dictionary analysis may have little impact on overall results when all the scores are equally biased, but it makes comparison of results from different dictionaries nearly impossible.

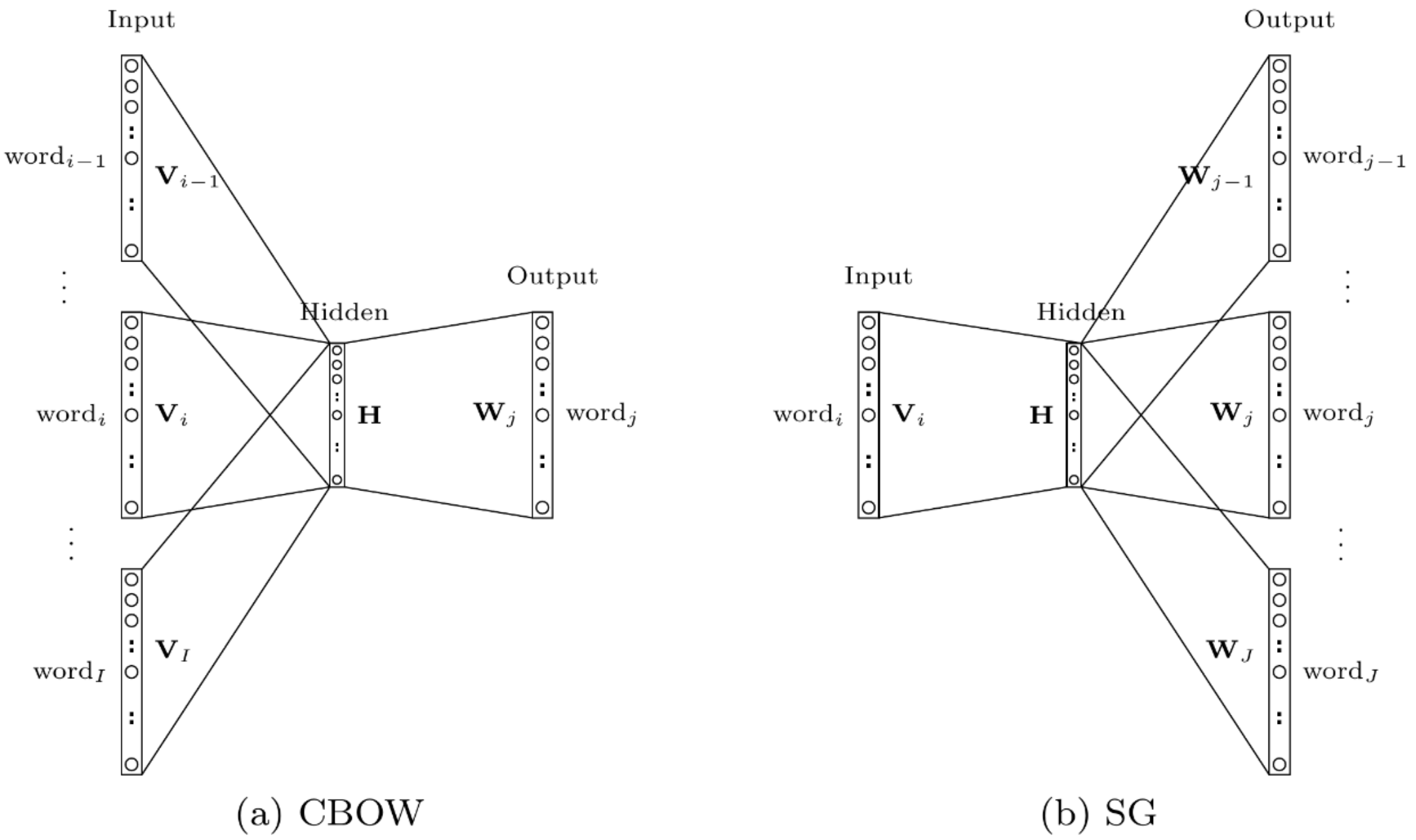

Figure1: Word2vec algorithms. $V$, $H$ and $W$ are matrices for the input layer values, hidden layer values and the output layer weights, respectively. For $K$-dimensional modes, $V_i = [v_1, v_2, \cdots, v_K]$, $H = [h_1, h_2, \cdots, h_K]$ and $W_i = [w_1, w_2, \cdots, w_K]$.

### Language Modeling

Word2vec is a type of masked language model that attempts to make best predictions of words hidden from the algorithm at the time of training using shallow neural networks consisting of the input, hidden and output layers. The model can be trained using either the skip-gram (SG) or the continuous bag-of-words (CBOW) algorithm (Figure 1).

The SG algorithm predicts each target word $j$ within $d$ words from the context word, while the CBOW algorithm makes prediction based on a group of words around the targe word $j$. To achieve this, the SG algorithm simply copies the input layer values $V$ of the context word $i$ to the hidden layer $H$ whereas the CBOW algorithm averages the input layer values of words occurring within the window $d$ from the target word $j$. The SG algorithm is simpler but its computational cost for training is significantly higher because it must predict words roughly $d$ times more.

SG:

$$H = V_i$$

CBOW:

$$H = \sum_{i \neq j}^{i \in I} \frac{V_i}{|I|} \text{ where } I = [j - d, \cdots, j + d\,]$$

With the hidden layer values $H$ and the output layer weights $W$, the algorithms can compute the probability of the occurrences of the target word $j$ given a context word $i$. The probability $q_i$ is delivered by applying the sigmoid (logistic) function to the sum of the products of the hidden layer values and the output layer weights for the target word.

$$r_{ij} = \sum_{k=1}^{K} H_k W_{jk}$$

$$q_{ij} = \text{sigmoid}(r_{ij})$$

$$= \frac{1}{1 + e^{-r_{ij}}}$$

The algorithms iteratively update the input layer values and the output layer weights using the back-propagation technique to achieve the highest prediction accuracy.[6] After training, spatial models use only $V$ as word vectors, but probabilistic models require both $V$ and $W$ to predict word occurrences.

---

[6] First, the output layer weights $W$ are updated based on the success or failure in predicting occurrences of words given the current hidden layer values $H$. Second, the input layer values $V$ are updated based on the errors caused by the hidden layer values.

**Polarity Estimation**

While the spatial model computes the word polarity scores $\hat{g}$ based on their average cosine similarity to seed words in $V$, the probabilistic model estimates the scores based on the average probability for the seed words to occur in the same context. The probability is delivered from $V$ and $W$ in the similar way as above training algorithm. $S = [s_1, s_2, \cdots, s_M]$ and $P = [p_1, p_2, \cdots, p_M]$ denote the seed words and their initial polarity, respectively, in both models.

Spatial model:

$$\hat{g}_i = \frac{1}{M} \sum_{m=1}^{M} \text{cosine}(V_i, V_m) p_m$$

Probabilistic model:

$$r_{im} = \sum_{k=1}^{K} V_{ik} W_{mk}$$

$$\hat{g}_i = \frac{1}{M} \sum_{m=1}^{M} \text{sigmoid}(r_{im}) p_m$$

Usually seed words receive the highest estimated polarity scores in spatial models, but they do not in probabilistic models due to the asymmetric relationship between context words and target words. This allows probabilistic models to assign extreme polarity scores to relevant words, making the result less sensitive to users' choice of seed words.

### Optimization

In addition to the greater interpretability and comparability of the probabilistic model, the uses of word2vec as a language model allows optimization of hyperparameters using the perplexity measure. When $\hat{q}_{dm}$ is the predicted probability of seed word $m$ in document $d$, the perplexity is the cross-entropy of their probability and observed frequency.[7]

$$\text{perpexity} = \exp\left( -\sum_{d=1}^{D}\sum_{m=1}^{M}\frac{f_{dm}}{N_d}\log(\hat{q}_{dm}) \Bigg/ \sum_{d=1}^{D}\sum_{m=1}^{M} f_{dm} \right)$$

## Evaluation

To demonstrate the advantages of the proposed technique, I evaluate the ability of the special and probabilistic models to replicate dictionary analysis simulating the absence of appropriate off-the-shelf dictionaries. In such situations, researchers can replace dictionary analysis with LSS. The goal is similar to synonym identification tasks in computer science (e.g., Arianto et al., 2024; Kakuta et al., 2025; Mikolov et al., 2013), but I evaluate estimated polarity of documents (instead of similarity of words) considering the purpose of LSS.

In this evaluation, I apply the models to a corpus of news articles published by China Daily between 2019 to 2023.[8] The analysis focuses on the English-language newspaper's emphasis on achievement in health issues during the COVID pandemic because its coverage is controlled by the Communist Party to promote positive image of China in overseas (Chen, 2012; Li & He, 2025). The evaluation follows five steps:

---

[7] The predicted probabilities $\hat{q}_d$ are normalized to ensure that they sum to 1.0.

[8] The corpus contains 91,709 articles downloaded through the Factiva API without any keyword filter.

1) I perform full dictionary analysis of the articles using LIWC (Pennebaker et al., 2015) to obtain the benchmark scores. LIWC contains keywords about ‘health’ (294 words) and ‘achievement’ (213 words), which were manually collected for social and psychological research.[9]
2) I randomly sample 10 sets of 10 unipolar seed words from the full dictionary for LSS. These seed words are not optimal but allow evaluation of models in situations where seed words chosen by users are relevant but not necessarily most extreme words.[10]
3) I apply spatial and probabilistic models to the same articles with different hyperparameters.[11] These models are trained using SVD or word2vec on the corpus with a sample of the unipolar seed words taken from the full dictionary. This evaluation results in sets of scores in 3,100 conditions in total.
4) I correlate the polarity scores of documents produced by the full dictionary and the three types of LSS models without any aggregation. In this comparison, I also included mini dictionaries that only comprise the seed words to highlight the contribution of the LSS models.
5) I analyze the relationship between the correlation coefficient and perplexity scores. If strong correlation is found between them, perplexity scores can be used to optimize hyperparameters such as the size of the input layer or the training algorithm in the absence of gold standard.

[9] The LWIC dictionaries are created in three steps: (1) collect candidate words from English dictionaries and thesauri, (2) select only relevant words from the candidate words by employing three independent judges, (3) remove words that occur infrequently or cause inconsistency by applying the dictionaries to example texts (Pennebaker et al., 2015).

[10] See Appendix for the list of seed words used in the evaluation.

[11] In word2vec, the windows size $d$ is 5 for CBOW and 10 for SG. The size of the input layer $k$ is 50, 100, 150, 200, 250 or 300; the learning rate is 0.05; and the number of iterations is 10 for both algorithms. In LSS, seed words are initialized with uniform polarity scores $p$ that sum to one.

**Results**

The results of evaluation show that word2vec-based probabilistic models are most strongly correlated with the full dictionaries in both ‘achievement’ and ‘health’ (Figure 2). The correlation is particularly strong when the SG algorithm is used to train them, reaching r=0.51 and r=0.66 in the respective concepts. The spatial models were also trained using the same algorithm, but they achieved significantly weaker correlations (r=0.47 and r=0.54, respectively).

The SVD-based models performed poorly, achieving only some of the lowest median scores (r= 0.34 and r=0.48). Their correlation is roughly the same as word2vec-based spatial models trained with the CBOW algorithms (r=0.36 and r= 0.40). They were even outperformed by the mini dictionaries (r=0.44) in ‘achievement’.

The variance of the correlation coefficients is the greatest in the mini dictionaries and the SVD-based models, followed by word2vec-based spatial models. Their large variance indicates that they are sensitive to the choice of seed words and hyperparameters. In contrast, the word2vec-based probabilistic models achieve the smallest variance in ‘health’, showing their high reliability.

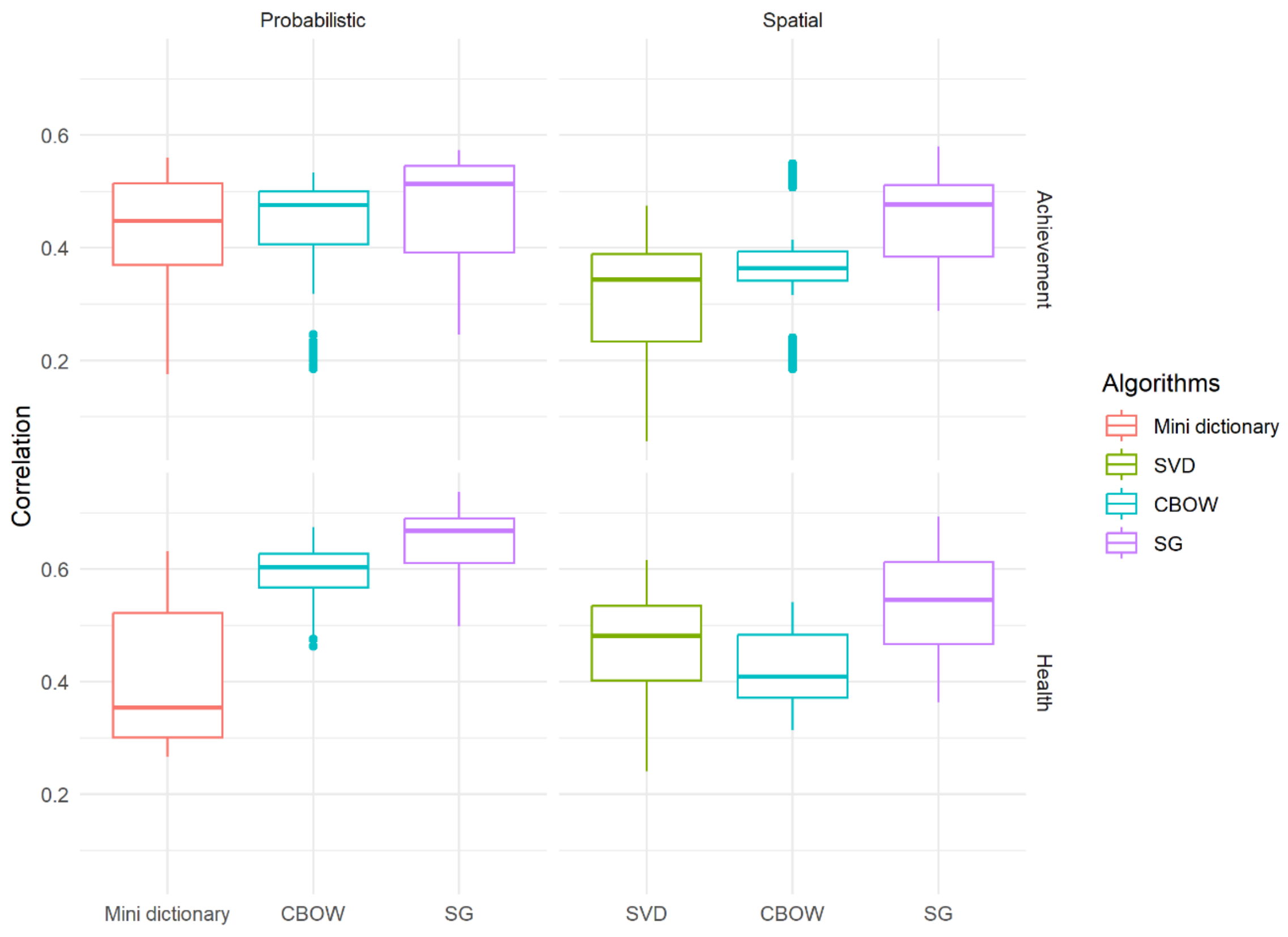

Figure 2: Correlation of document polarity scores. The vertical axis is Pearson's correlation coefficient. The boxes show the quantile range of the correlation coefficients.

The perplexity scores of the probabilistic models tend to be smaller in 'health' than in 'achievement' (Figure 3), corresponding to the larger correlation coefficients in the former than in the latter. Within the concepts, strongly negative association can be also found between the perplexity scores and the correlation coefficients for the same seed words. The perplexity scores are significantly smaller for models trained using the SG algorithm with only two exceptions in 'achievement.' The size of the input layer caused only small variation in both perplexity scores and correlation coefficients.

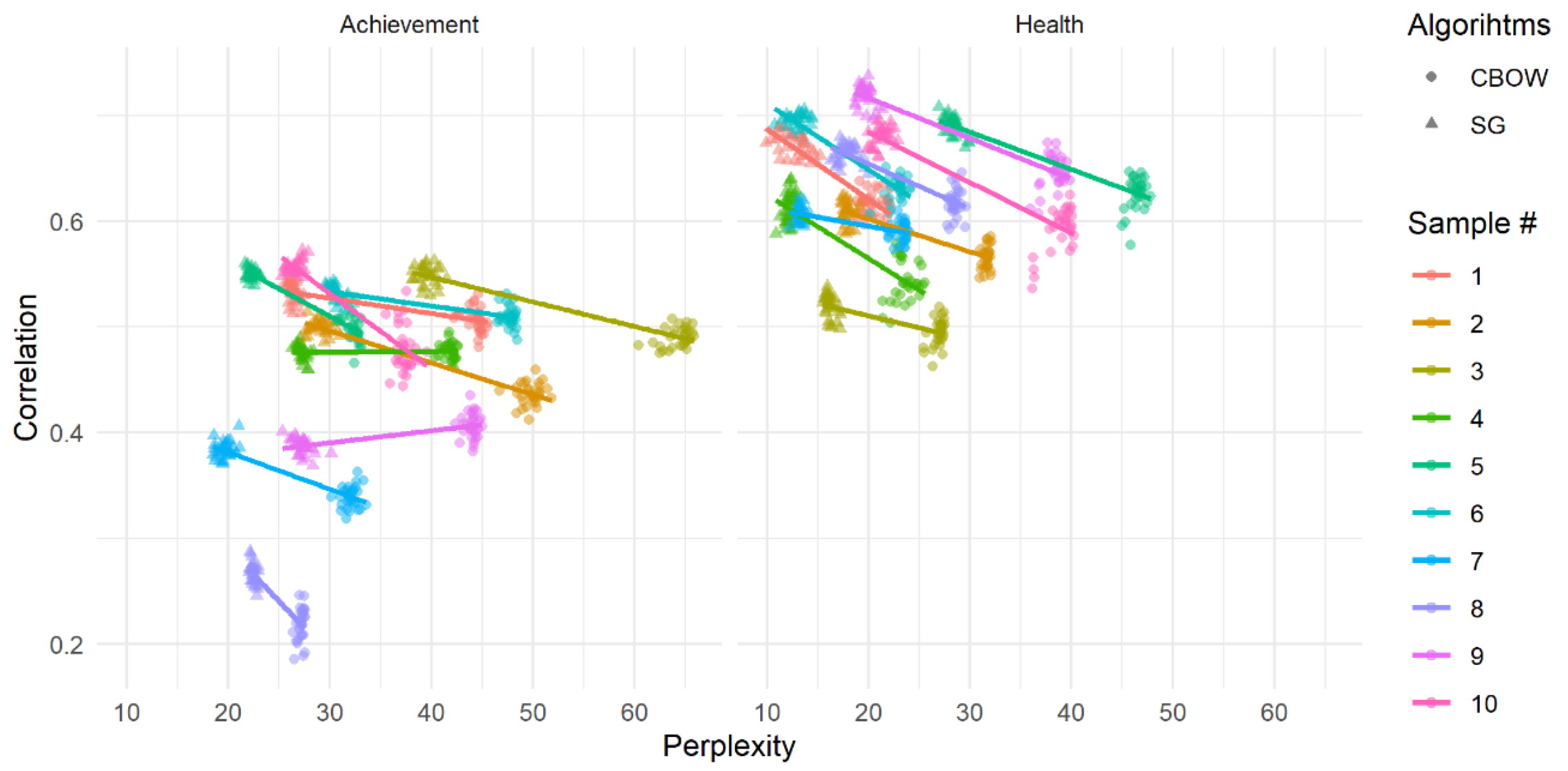

Figure 3: Perplexity of probabilistic models. Colors indicate 10 samples of seed words; horizontal axis is perplexity scores for seed words; vertical axis is correlation coefficients. Models were trained with 10 different samples of seed words.

## Example

I selected a model for each concept that achieved the lowest perplexity scores in above evaluation: the model for 'achievement' is 150-dimensional and seeded with sample 7 (perplexity=18.47); the model for 'health' is 200-dimensional and seeded with sample 1 (perplexity=9.87); both models were trained using the SG algorithm. The correlations of their document polarity scores with the full dictionaries were modest (r=0.37 and r=0.67 in 'achievement' and 'health', respectively) but their daily average scores are very similar (r=0.87 and r=0.92, respectively).

Figures 4 and 5 show the estimated polarity of words by the chosen models. In 'achievement', the full dictionary words and seed words are largely overlapped with other words. This explains the weak correlation of the document scores produced by the full dictionary and the model. Despite the model's poor performance, it achieved the lowest perplexity scores due to the very frequent seed words such as "opportunities" and "promote".

In contrast, the full dictionary words and the seed words for 'health' are distributed distinctively from other words. This suggests that the choice of seed words is close to optimal, enabling the algorithm to assign high polarity scores to non-seed words. These seed words also allowed the selection of one of the best-performing models based on the perplexity score.

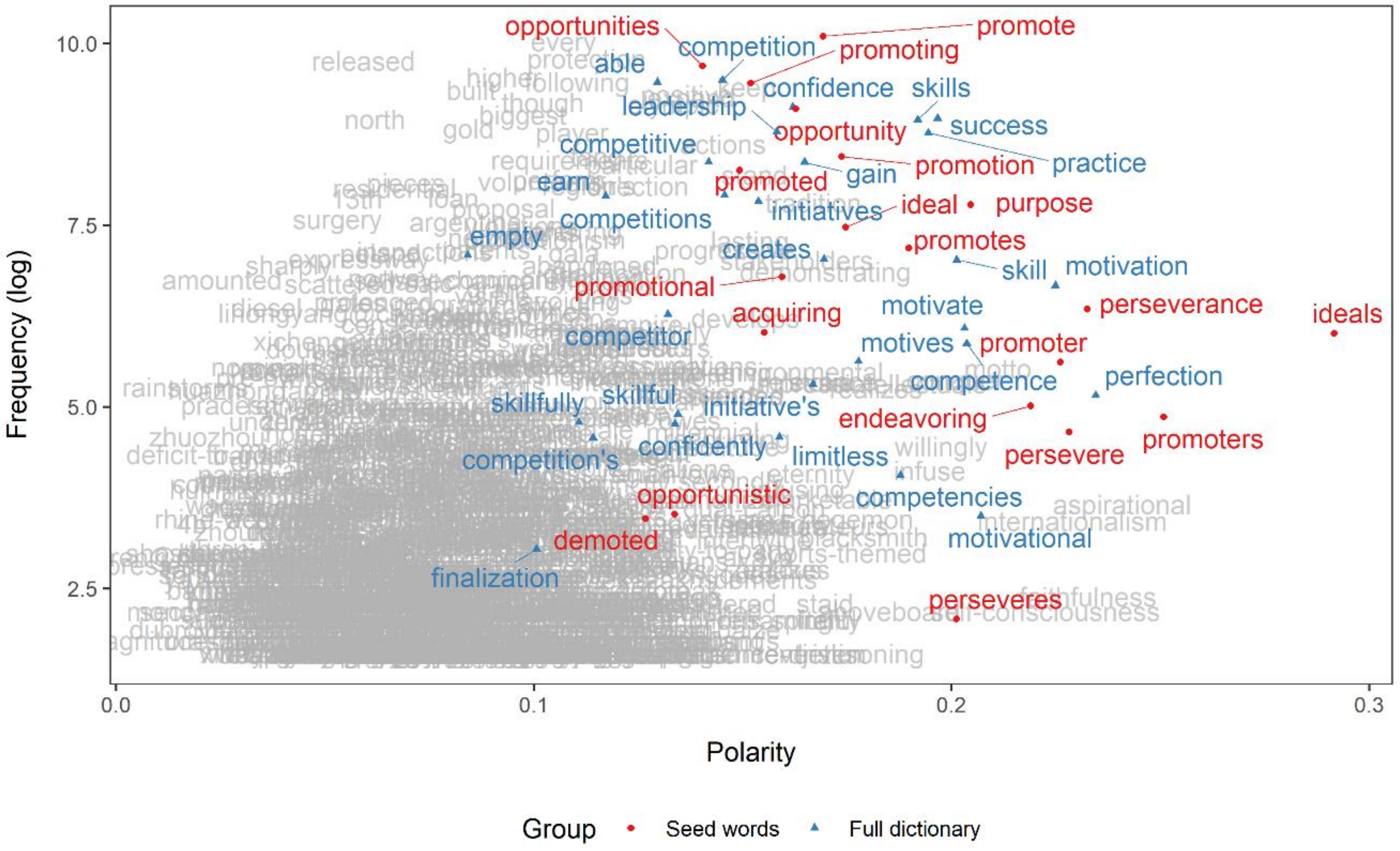

Figure 4: Polarity word for 'achievement'. The horizontal axis is the polarity scores; the vertical axis is the frequency in the corpus. The seed words and full-dictionary words are highlighted in red and blue, respectively.

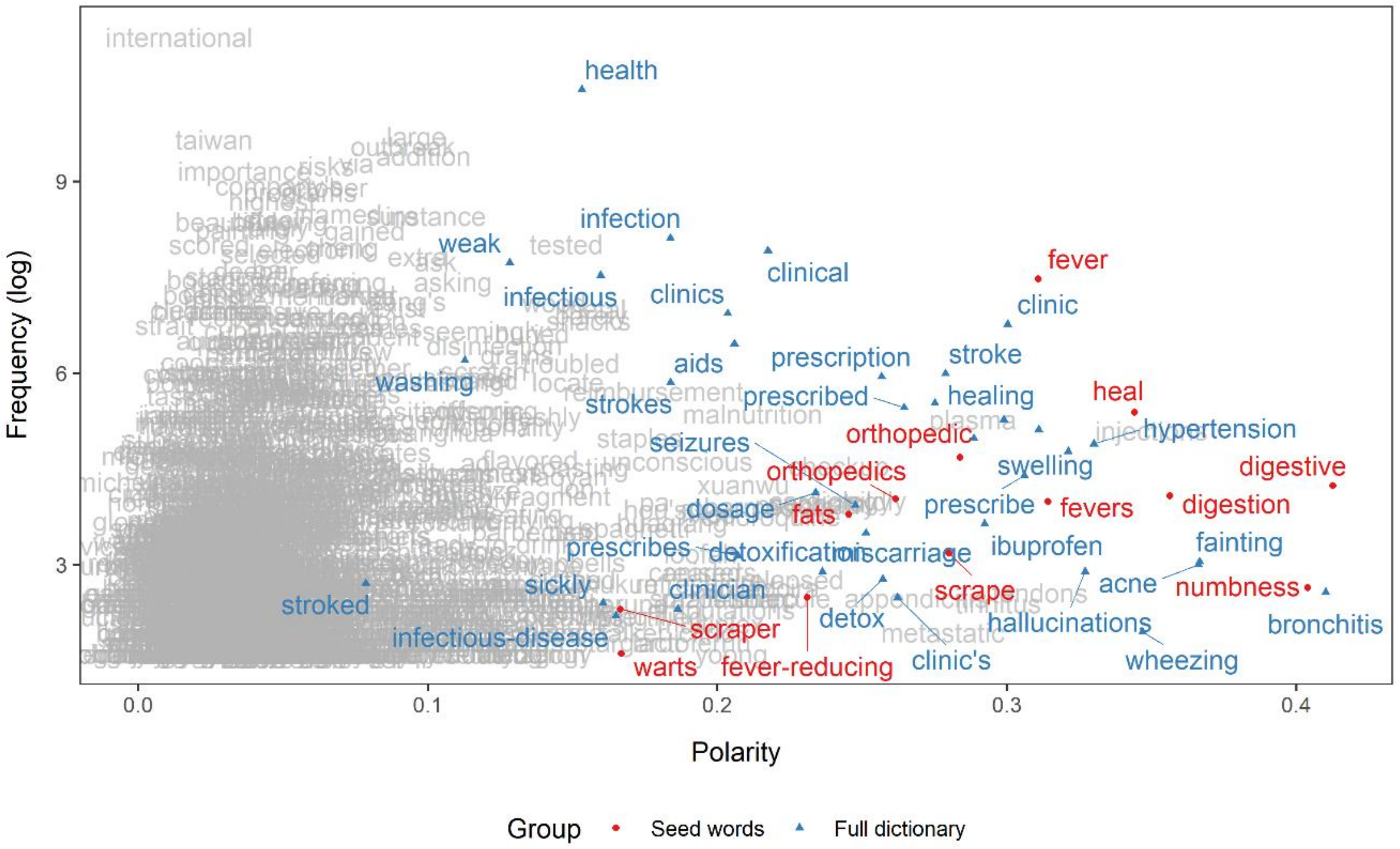

Figure 5: Polarity words for 'health'. The horizontal axis is the polarity scores; the vertical axis is the frequency in the corpus. The seed words and full-dictionary words are highlighted in red and blue, respectively.

To reveal China Daily's coverage during the COVID crisis, I smoothed document polarity scores separately for the articles about China or other countries (Figure 6).[12] The top-left plot shows that the newspaper's emphasis on China's achievement weakened in the beginning of crisis, but it became stronger gradually towards the end of 2023, reaching 10% more likely for the seed words to appear in articles about China. The top-right plot shows that the newspaper focused on health issues in China in the beginning of 2020 when the crisis begun, but its focus shifted to other countries; the coverage of health issues decreased gradually towards 2023.

[12] In this example, I classified articles only using the name of the country and the capital city (China, Chinese, Beijing), but it is desirable to include more placenames in substantive research.

To understand the newspaper's emphasis on China's achievement in health issues, I combined the two sets of polarity scores by taking the square root of the product in each document. The combined scores can increase only when the newspaper emphasizes achievement in articles about health issues. Unlike the original scores, the combined scores for China become significantly higher than other countries only during limited periods, two of which are the beginning and the end of China's zero-COVID policy. The sudden end of the lockdown in Chinese cities in late 2022 was followed by the state-owned media's effort to frame the strict public health policy as success (Nordin, 2023).

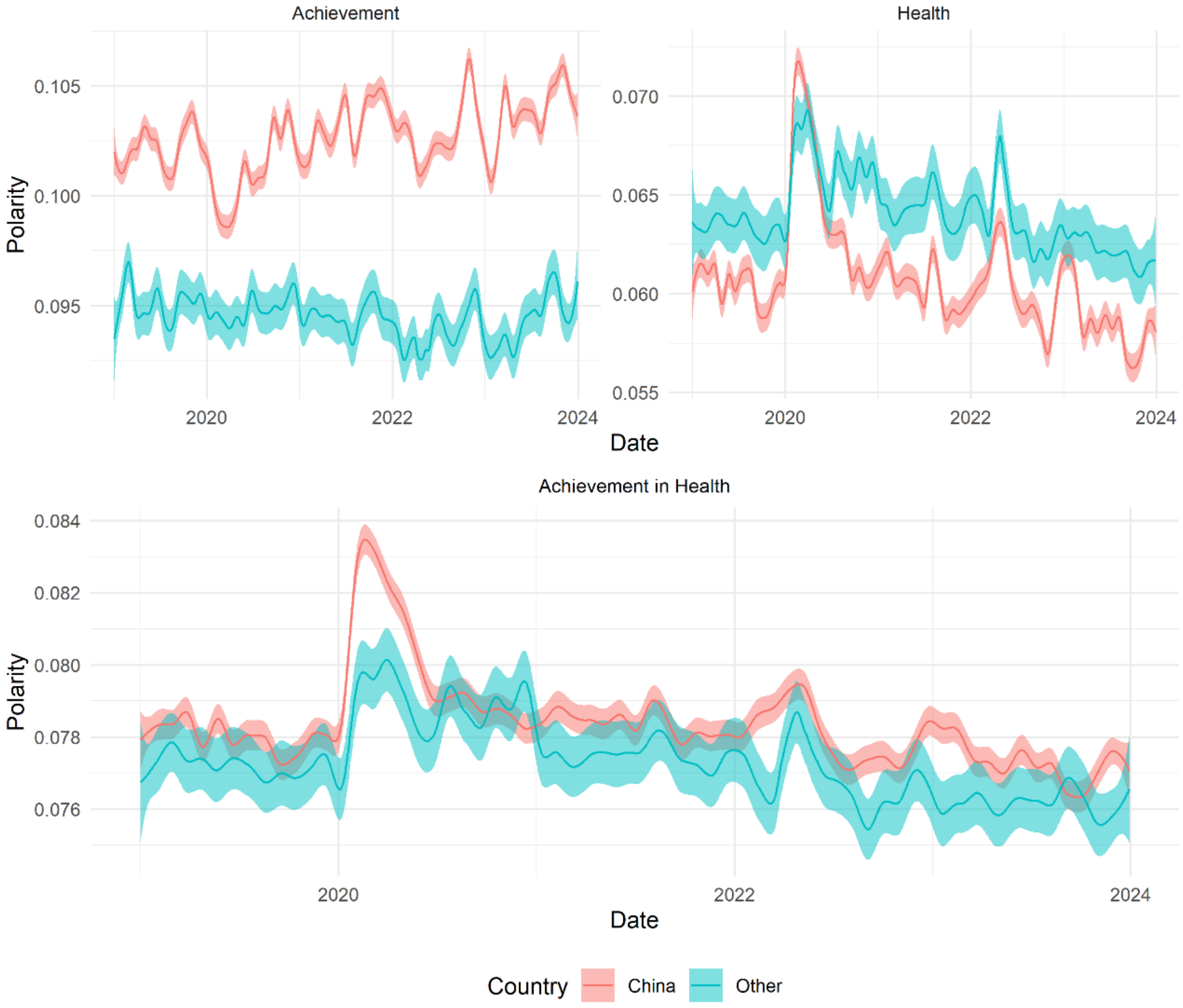

Figure 6: Document polarity in China's Daily. The plots on top are document polarity scores for 'achievement' and 'health'. The plot on the bottom is the combined document polarity scores for 'achievement in health'. Red and blue lines are the polarity scores of documents about China and other countries, respectively. Bands around the lines are the 95% confidence intervals of local regression smoothing.

## Conclusions

The systematic comparison revealed that word2vec improves both spatial and probabilistic LSS models, but the changes were more dramatic in the latter. Word2vec allowed the probabilistic models to accomplish the strongest correlation with the full dictionaries, while the spatial models achieved only the lowest correlation even being outperformed by mini dictionaries in one concept.

In both spatial and probabilistic models, the correlation with the full dictionaries was higher in the SG algorithm, but the difference between the training algorithms was significantly smaller in probabilistic models. This suggests that the benefit of the probabilistic models is even greater when computing resources are limited because they can be trained reasonably well using the less expensive CBOW algorithm.

The greater performance of the probabilistic models is explained by their ability to assign large polarity scores to words in the corpus. Although the random sampling of seed words did not guarantee that they are the most extreme words, the probabilistic models produced results similar to the full dictionaries. This makes choice of seed words easier for users and the results of analysis less sensitive to their choice.

The probabilistic models also allow users to optimize the hyperparameters based on their goodness-of-fit to the data. In the evaluation, the perplexity scores correctly indicated that the SG algorithm is more appropriate than the CBOW algorithm in most cases, although the scores were only minimally affected by the input layer sizes. The models' insensitivity to the input layer sizes is welcoming because users can choose values without the risk of over or underfitting.

The choice of seed words appears to be easier for the probabilistic models, but they still need to be highly relevant to the target concept. In choosing seed words, users should manually inspect words that receive high polarity scores in the models.[13] If the words are often used to discuss the targe concept, the choice of seed word is appropriate. Yet users should avoid selecting seed words based on the perplexity scores because above optimization method relies on the existence of appropriate seed words.

In the example, the probabilistic models revealed the tendency of the China Daily's coverage very clearly. The difference between articles about China and other countries were significant nearly throughout the period, but more nuanced pictures emerged when the polarity scores for the two concepts interacted. This is an additional benefit of probabilistic polarity scores that range only between 0 and 1.

Finally, this study demonstrated that masked language models have great potential for semisupervised text analysis because they can predict occurrences of seed words in given contexts. Since increasing numbers of neural network models are being developed, it should be possible to further enhance semisupervised techniques to capture more complex and unique ideas in texts.

[13] Use can generate a list of context words that have high probabilities for each seed word in the underlying model using the probability() function in the wordvector package.

## Appendix

Seed words are sampled randomly from the LIWC dictionaries (Pennebaker et al., 2015) in the evaluation. Words such as "inadequa*" and "incapab*" may not be appropriate for measuring emphasis on achievement but are included in the samples.

### Achievement

| Sample # | Words |
|---:|---|
| **1** | compet*, skill*, proudly, ideal*, goal*, dropout*, improving, victor*, resolv* |
| **2** | persever*, opportun*, goal*, limit*, elit*, motiv*, ideal*, trying, reward*, promot* |
| **3** | superb*, opportun*, initiat*, domina*, surviv*, champ*, compet*, advanc*, prize*, goal* |
| **4** | skill*, trying, beaten, medal*, improving, incapab*, persever*, award*, winn*, compet* |
| **5** | emptiness, excellent, beaten, skill*, overconfiden*, obtain, strateg*, compet*, succeed*, progress |
| **6** | accomplish*, lead, winn*, incompeten*, compet*, efficien*, champ*, honor*, obtain, conquer* |
| **7** | ideal*, promot*, persever*, purpose*, wins, endeav*, opportun*, demot*, acquir* |
| **8** | inadequa*, domina*, unproduc*, achievable, fail*, gain*, obtains, celebrat*, losing |
| **9** | dropout*, team*, goal*, overtak*, victor*, abilit*, skill*, successes, surviv*, surpass* |
| **10** | recover*, initiat*, medal*, promot*, practicing, triumph*, practices, persist*, skill*, leads |

### Health

| Sample # | Words |
|---:|---|
| **1** | warts, coronar*, orthoped*, numbness, heal, fats, digest*, healthily, scrape*, fever* |
| **2** | lump*, injur*, fatter, immun*, doctor*, headache*, weakest, rehab*, thermometer* |
| **3** | workout*, tox*, physical, yawn*, injur*, ibuprofen, digest*, wash, poison*, clinic* |
| **4** | optometr*, rehab*, cancer*, weak, inflam*, workout*, healthily, hiccup*, lives, blind* |
| **5** | workout*, doctor*, drug*, cancer*, operat*, iv, injur*, patholog*, contag*, pharmac* |
| **6** | addict*, illness*, withdrawal, strept*, hormone*, tenderly, cough*, disease*, testosterone*, chronic* |
| **7** | headache*, neurolog*, living, nurse*, stroke*, drows*, syndrome*, injur*, diagnos* |
| **8** | hemor*, digest*, rehab*, addict*, syphili*, washes, drug*, exhaust*, weakest |
| **9** | digest*, prescri*, faint*, livel*, schizophren*, ache*, tox*, orthoped*, pharmac*, pimple* |
| **10** | operat*, tox*, ibuprofen, infertil*, nause*, medic*, rash*, lymph*, pain |